\documentclass[conference]{IEEEtran}
\IEEEoverridecommandlockouts

\usepackage{diagbox}
\usepackage{array}
\usepackage{booktabs}
\usepackage{cite}
\usepackage{amsmath,amssymb,amsfonts}
\usepackage{algorithmic}
\usepackage{graphicx}
\usepackage{textcomp}
\usepackage[most]{tcolorbox}
\usepackage{xcolor}
\usepackage{orcidlink}
\usepackage[linesnumbered, vlined, ruled]{algorithm2e}
\SetAlFnt{\footnotesize}
\SetAlCapFnt{\footnotesize}

\SetCommentSty{mycommfont}
\usepackage{wrapfig}
\usepackage{booktabs}
\usepackage{multirow}
\usepackage{array}
\usepackage{hyperref}
\usepackage{subcaption}
\def\BibTeX{{\rm B\kern-.05em{\sc i\kern-.025em b}\kern-.08em
    T\kern-.1667em\lower.7ex\hbox{E}\kern-.125emX}}

\newtcolorbox{highlightbox}{
  colback=yellow!20,
  colframe=yellow!80!black,
  boxrule=0.5mm,
  arc=3mm,
  auto outer arc,
  left=1mm, right=1mm, top=1mm, bottom=1mm
}

\begin{document}

\title{
Enhancing lidar Point Cloud Sampling via Colorization and Super-Resolution of lidar Imagery\\
}


\author{
    \IEEEauthorblockN{
        \vspace{1em}
        Sier Ha\,\orcidlink{0009-0000-3617-107X}\,
        Honghao Du\,\orcidlink{0009-0008-7600-0302}\,
        Xianjia Yu\,\orcidlink{0000-0002-9042-3730}, 
        Tomi Westerlund\,\orcidlink{0000-0002-1793-2694}
    }
    \IEEEauthorblockA{
        \normalsize
        \href{https://tiers.utu.fi}{Turku Intelligent Embedded and Robotic Systems (TIERS) Lab, University of Turku, Finland}.\\
        Emails: \{sier.s.ha, honghao.h.du, xianjia.yu, tovewe\}@utu.fi\\[+6pt]
    }
}

\maketitle

\begin{abstract}
    Recent advancements in lidar technology have led to improved point cloud resolution as well as the generation of  360\textdegree\, low-resolution images by encoding depth, reflectivity, or near-infrared light within each pixel. 
    These images enable the application of deep learning (DL) approaches, originally developed for RGB images from cameras to lidar-only systems, eliminating other efforts, such as lidar-camera calibration. Compared with conventional RGB images, lidar imagery demonstrates greater robustness in adverse environmental conditions, such as low light and foggy weather. 
    Moreover, the imaging capability addresses the challenges in environments where the geometric information in point clouds may be degraded, such as long corridors, and dense point clouds may be misleading, potentially leading to drift errors.
    
    Therefore, this paper proposes a novel framework that leverages DL-based colorization and super-resolution techniques on lidar imagery to extract reliable samples from lidar point clouds for odometry estimation.
    The enhanced lidar images, enriched with additional information, facilitate improved keypoint detection, which is subsequently employed for more effective point cloud downsampling.
    The proposed method enhances point cloud registration accuracy and mitigates mismatches arising from insufficient geometric information or misleading extra points. Experimental results indicate that our approach surpasses previous methods, achieving lower translation and rotation errors while using fewer points.
\end{abstract}



\section{Introduction}\label{sec:introduction}
lidar sensors have become increasingly significant in various domains of robotics and autonomous systems, particularly in navigation, perception, lidar Odometry (LO)~\cite{zhang2014loam,zhang2017low}, and Simultaneous Localization and Mapping (SLAM)~\cite{jiang2022autonomous, sier2023benchmark}. Key factors that facilitate lidars' utility are the progressively increasing precision and density of point cloud data, which offers extensive geometric information about the surroundings. 
However, when calculating accurate LO or SLAM, the dense point cloud and conventional sampling approaches may introduce more inaccuracies, leading to error drift. This issue becomes particularly evident in environments where geometric information is degraded, such as tunnels and corridors~\cite{pfreundschuh2024coin}. Consequently, the process of extracting relevant points from the lidar point cloud is of notable importance for effective point cloud registration.

\begin{figure*}[t]
    \centering
    \includegraphics[width=0.98\textwidth]{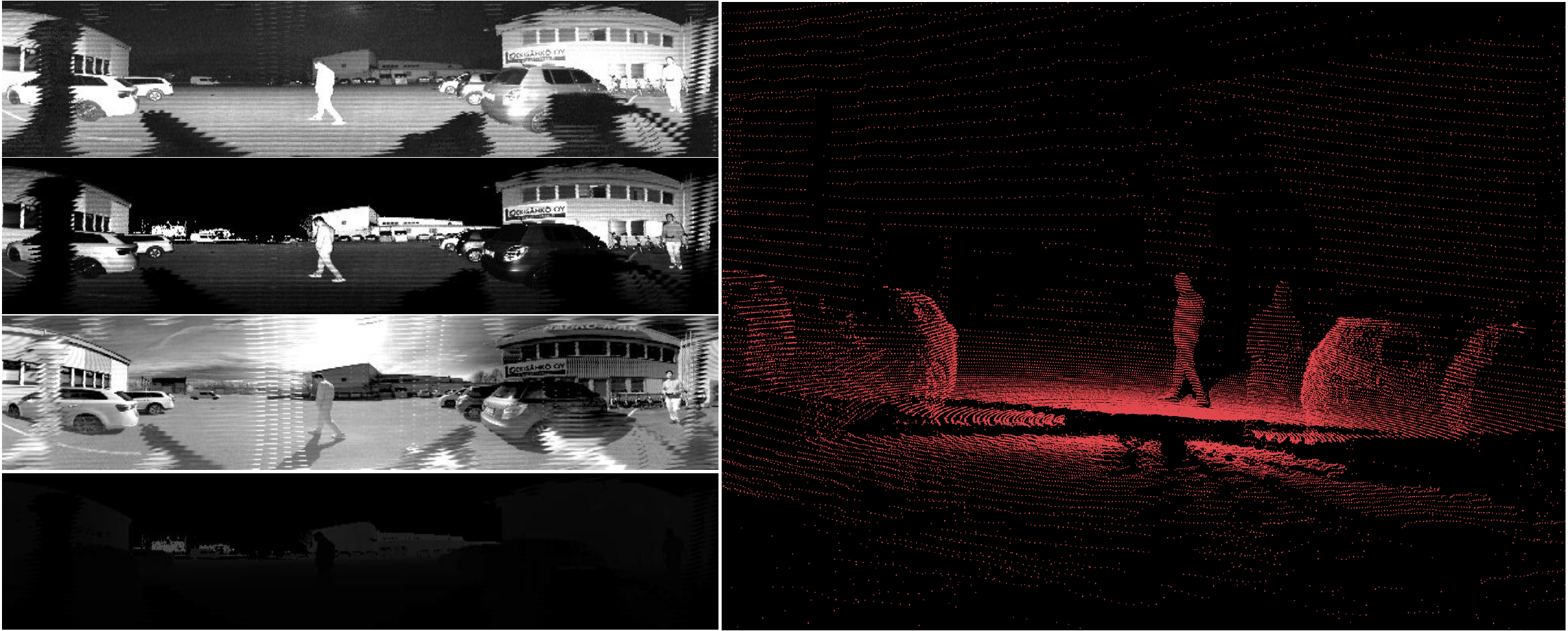}
    \caption{Visualization of Ouster lidar data. From top to bottom on the left: signal intensity, reflectivity, near-infrared (near-IR), and depth image. The right side displays the corresponding point cloud.}
    \label{fig:ouster-lidar-img}
    \vspace{-0.5em}
\end{figure*}

Nowadays, more lidar manufacturers, are increasingly focusing on enhancing their devices' capabilities to directly produce imaging data, moving beyond traditional point cloud outputs. Particularly, lidars, such as Ouster lidar~\cite{tampuu2022lidar,angus2018lidar,xianjia2022analyzing}, utilize specialized photon-counting hardware and large-aperture optics to inherently generate fixed-resolution images encompassing intensity, ambient near-infrared illumination, and depth—all perfectly spatially correlated and free from temporal mismatch or shutter-related artifacts. These image data, captured entirely by the lidar without additional sensors, closely mimic conventional photographic images, facilitating seamless application and adaptation of existing deep learning (DL) algorithms originally developed for cameras. Consequently, this integrated approach eliminates the need for lidar-camera calibration and avoids the inherent inaccuracies associated with sensor fusion, thus significantly streamlining the deployment of DL solutions in lidar-based perception tasks~\cite{yu2023general}. Even though these lidar images are more robust to adverse environments like varying lighting conditions and foggy weather, they are often low resolution.
 An example of Ouster lidar imageries is illustrated in Fig.~\ref{fig:ouster-lidar-img}.  

\begin{figure}[t]
    \centering
    \includegraphics[width=0.49\textwidth]{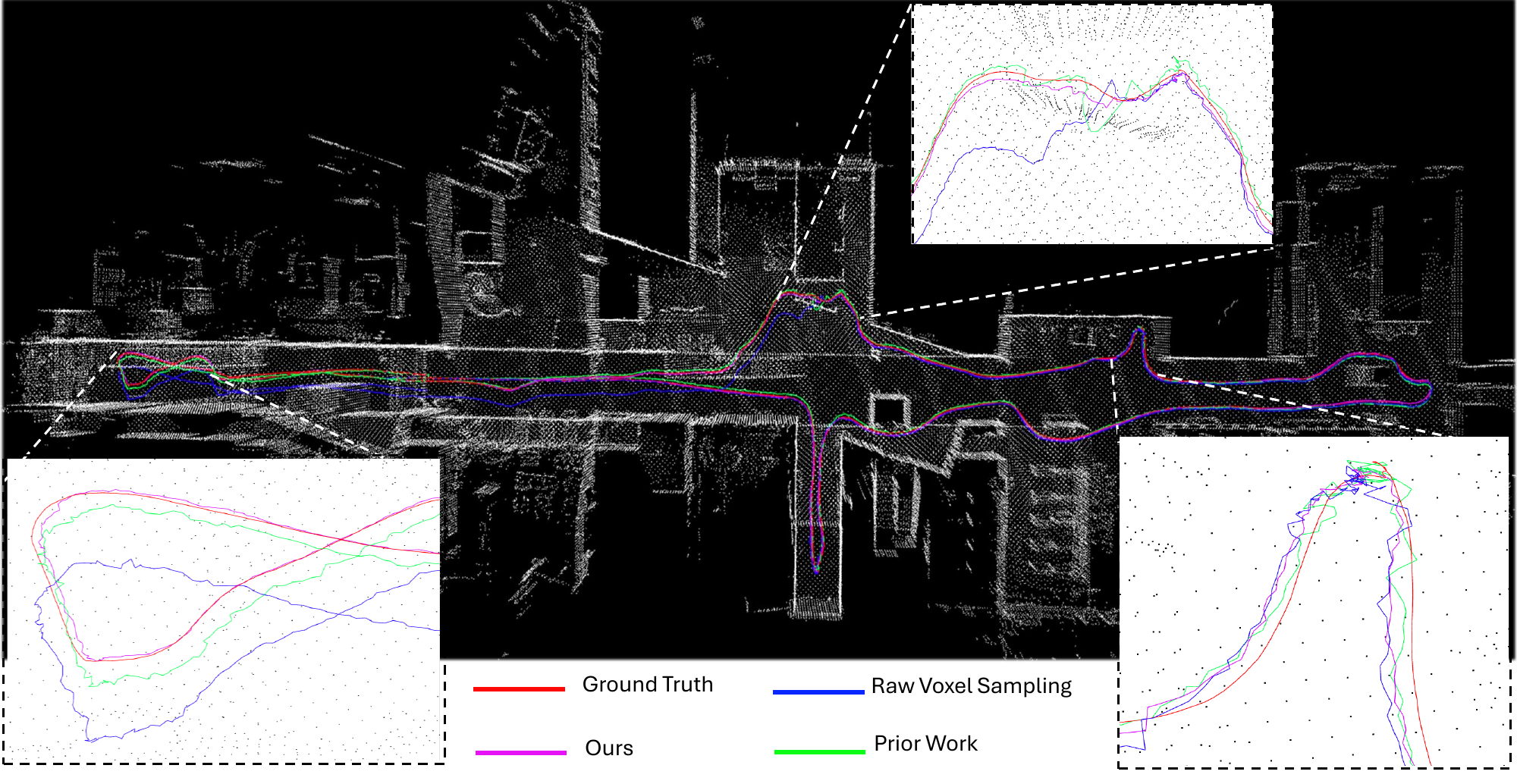}
    \caption{An illustrative example of potential drift and the effectiveness of the proposed approach is presented. In particular, the trajectories of two corners (in the bottom-left and top-right zoomed boxes) show that our approach aligns more closely with the ground truth while utilizing fewer points.}
    \label{fig:drift-example}
    \vspace{-0.5em}
\end{figure}

In the context of enhancing the accuracy and robustness of LO, existing research has primarily focused on traditional methods utilizing intensity information for the displacement calculation between image frames~\cite{pfreundschuh2024coin ,dong2023r}. 
In our previous work, we assessed the performance of keypoint detectors and extractors on lidar imageries, exploring their potential to improve the robustness of LO~\cite{zhang2023lidar}. However, there remains considerable scope for leveraging other DL approaches such as DL-based colorization and super-resolution to further enhance system robustness. 
Given that lidar imageries are low-resolution, these techniques demonstrate significant potential for enhancement. 
Additionally, unlike in our previous work, we employed key point extractors across all color channels (R, G, and B) of the image, rather than restricting the extraction to a single channel.


To address these issues, building upon the comprehensive review and detailed comparison of existing DL-based super-resolution and colorization methods for lidar imageries presented in our previous research~\cite{ha2024enhancing}, we propose enhanced techniques for keypoint extraction. We applied these DL-based methods to various combinations of lidar imageries and evaluated their effectiveness in the lidar odometry (LO) task. Our results indicate significant improvements in accuracy and robustness compared to our prior approaches. Specifically, Fig.~\ref{fig:drift-example} presents a comparative analysis demonstrating enhanced performance over previous methods, including scenarios involving no downsampling, voxel-based downsampling, and baseline approaches.

The remainder of this paper is structured as follows. Section~\ref{sec:related_work} provides the foundational background for this study, covering point cloud sampling methods, keypoint extraction techniques, and deep learning-based colorization and super-resolution approaches. Section~\ref{sec:METHODOLOGY} details the overall workflow and evaluation scheme of the proposed framework. The effectiveness of the proposed method is demonstrated through experimental results in Section~\ref{sec:expert_resu}, followed by conclusions and directions for future research in Section~\ref{sec:conclusion}.



\begin{table*}[t]
\centering
\resizebox{\textwidth}{!}{%
\begin{tabular}{l|ccccccccc}
\toprule
\diagbox{\textbf{sensor}}{\textbf{specification}}
 &
  \textbf{IMU} &
  \textbf{Type} &
  \textbf{Channels} &
  \textbf{Image Resolution} & 
    \textbf{FoV} &
  \textbf{Angular Resolution} &
  \textbf{Range} &
  \textbf{Freq} &
  \textbf{Points} \\
   \midrule
\textbf{Ouster OS0-64} &
  ICM-20948 &
  spinning &
  128 &
  $ 1024 \times 128 $ &
  $360^\circ \times 90^\circ$ &
  $V:0.7^\circ, H:0.18^\circ$ &
  50\,m &
  10\,Hz &
  2,621,440 pts/s \\ \bottomrule
\end{tabular}%
}

\caption{Specifications of Ouster OS0-128.}
\label{tabs:ouster-table}
\vspace{-0.5em}
\end{table*}

\section{Related Work} \label{sec:related_work}

This section presents the recent state-of-the-art work of point cloud sampling, keypoint extractors, and DL-based colorization and super-resolution.

\subsection{Point Cloud Sampling}
Point cloud sampling is a critical step in 3D data processing, particularly for efficiently handling large datasets generated by modern 3D scanning technologies. Non-learning-based methods like voxel downsampling and Farthest Point Sampling (FPS) are widely used. Voxel downsampling is common in applications such as lidar odometry~\cite{vizzo2023kiss,xu2021fast} and SLAM~\cite{ye2019tightly,wang2021f}, where it reduces point cloud size by replacing points within each voxel with a single representative point, though it may sacrifice fine details. FPS~\cite{eldar1997farthest}, often employed in DL applications, iteratively selects well-distributed points across the cloud, balancing uniform coverage with feature preservation~\cite{ruizhongtai2023deep}. 
However, these methods are not effective when there is a degradation of geometric information in the point cloud.

Recent advancements in DL have introduced methods like S-NET~\cite{dovrat2019learning} and PST-NET~\cite{wang2021pst}, which optimize sampling for specific tasks. S-NET learns task-specific sampling strategies, outperforming traditional methods like FPS by tailoring the selection of points to the needs of applications such as classification and retrieval. PST-NET further innovates by leveraging a point-based transformer to consider geometric relationships among points, incorporating features like self-attention and local feature extraction to generate an optimal resampling distribution. 

Despite these advancements, research in point cloud processing has predominantly focused on tasks like classification, segmentation, and object detection~\cite{qi2017pointnet,li2018so,qi2019deep}. Sampling strategies themselves have received relatively less attention.  This gap highlights the critical need for continued research into efficient and effective sampling techniques, particularly within learning-based frameworks, to further enhance performance across various point cloud applications.

\subsection{Keypoint Extractor}
Keypoint extraction is essential in computer vision, identifying salient points in an image that remain invariant to transformations like rotation, scaling, and illumination. Numerous methods have been proposed over time.
Scale-Invariant Feature Transform (SIFT)~\cite{lowe2004distinctive} uses a Difference-of-Gaussian (DoG) method to detect key points and compute invariant descriptors to image scaling and rotation. It is highly accurate but computationally expensive. Speeded-Up Robust Features (SURF)~\cite{bay2006surf} is a faster alternative to SIFT and uses box filters to approximate DoGs, thus enabling faster computation while maintaining robust performance. Features from Accelerated Segment Test (FAST)~\cite{rosten2006machine} is an efficient keypoint detector. It identifies key points by comparing the intensity of a pixel to the intensity of pixels in a circular neighborhood. FAST is designed for speed and is well-suited for applications requiring real-time processing. However, it does not provide orientation information, making it less robust to rotation. Binary Robust Independent Elementary Features (BRIEF)~\cite{calonder2010brief} constructs binary strings by comparing the intensities of random pixel pairs within smooth image blocks. Although BRIEF is extremely fast and memory efficient, it is not inherently rotation-invariant, which limits its robustness under rotation. Oriented FAST and Rotated BRIEF (ORB)~\cite{rublee2011orb} enhances the FAST detector by adding orientation information and coupling it to the BRIEF descriptor.

SuperPoint~\cite{detone2018superpointselfsupervisedpointdetection} is an end-to-end self-supervised neural network designed for feature detection and description in computer vision. It addresses the challenge of detecting and describing keypoints in images by learning from data through self-supervision. In many cases, SuperPoint outperforms traditional methods like SIFT and ORB. With its learned feature points and descriptors, the model excels in various computer vision tasks.

Accurate and Lightweight Keypoint Detection and Descriptor Extraction (ALIKE)~\cite{Zhao2023ALIKED} stands out from the crowd of methods because of its focus on both accuracy and computational efficiency. 
ALIKE implements a hybrid approach that utilizes classical computer vision techniques augmented with modern machine learning methods, and it enables accurate and reliable keypoint detection in a wide range of image conditions. The descriptors in ALIKE are designed to be unique and compact, less susceptible to image variations such as noise, lighting changes, or occlusion. It improves the high robustness against image transformations.




\subsection{DL-based Colorization and Super Resolution}\label{sec:dl-col-sr}
In our previous research~\cite{ha2024enhancing}, we presented a comprehensive overview of deep learning (DL)-based super-resolution and colorization methods tailored specifically for lidar-generated imagery. Existing DL approaches for colorization primarily include GAN-based frameworks (e.g., DeOldify~\cite{antic2019deoldify}, PearlGAN~\cite{luo2022thermal}, ChromaGAN~\cite{vitoria2020chromagan}), CNN-based architectures (e.g., DDColor~\cite{kang2023ddcolor}, DISCO~\cite{XiaHWW22}, InstColorization~\cite{Su-CVPR-2020}, Colorful Image Colorization~\cite{zhang2016colorful}), and diffusion models, each with unique strengths and limitations regarding visual realism, specific application scenarios, and computational complexity. Notably, DeOldify~\cite{antic2019deoldify} demonstrated strong performance in natural landscapes, while PearlGAN~\cite{luo2022thermal} and I2V-GAN~\cite{I2V-GAN2021} effectively converted infrared images into visible spectrum equivalents, despite their inherent domain and computational limitations. Regarding image super-resolution, methods evolved from early CNN-based models such as SRCNN~\cite{dong2015image} and VDSR~\cite{kim2016accurate} to advanced GAN-based techniques like SRGAN~\cite{ledig2017photo} and ESRGAN~\cite{wang2018esrgan}, and more recently to transformer-based architectures such as SwinIR~\cite{liang2021swinir} and CAT~\cite{chen2022cross}, progressively enhancing the reconstruction of fine details and realistic textures, though accompanied by increased computational demands. Comparative evaluations indicated that the CARN~\cite{ahn2018fast} model offers an optimal balance between image quality and computational efficiency, suitable for practical deployment. Additionally, DeOldify~\cite{antic2019deoldify} was identified as consistently effective for lidar-image colorization tasks with relatively low computational overhead. Collectively, these DL-based enhancement techniques significantly improve lidar image quality and interpretability, benefiting subsequent robotic applications including odometry estimation and 3D reconstruction.

\begin{figure*}[t]
    \centering
    \includegraphics[width=0.98\textwidth]{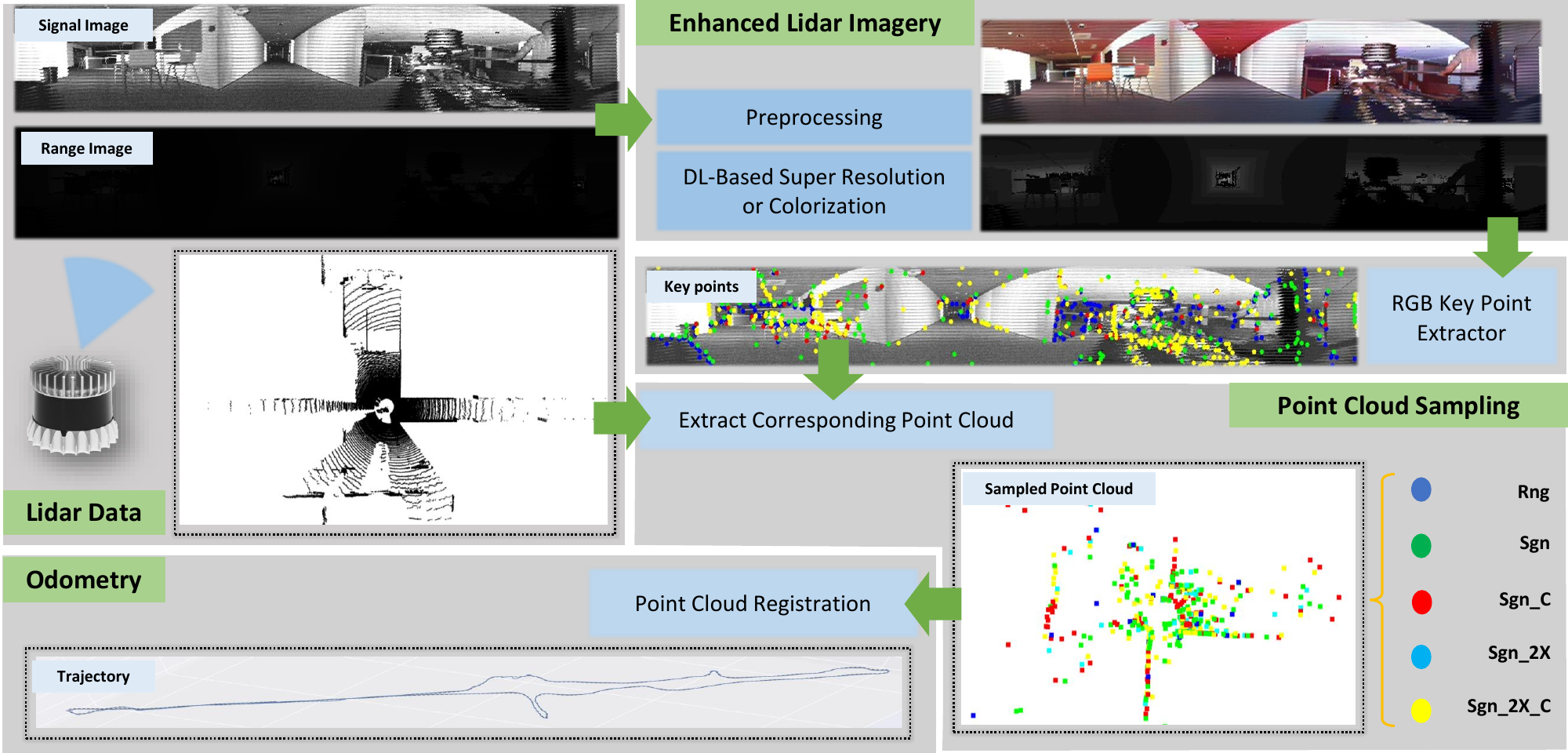}
    \caption{The system overview of the proposed approach. \textit{Rng} and \textit{Sng} denote the range and signal images, respectively. The subscript \textit{\_C} indicates that colorization has been applied to the corresponding image type, while \textit{\_2X} signifies the application of 2× super-resolution enhancement.}
    \label{fig:overview}
    \vspace{-0.5em}
\end{figure*}

\section{Methodology}\label{sec:METHODOLOGY}
\subsection{Dataset for Evaluation}\label{subsec:my_dataset}
We carry out all evaluations in the experiment with the published open-source multi-modal lidar datasets~\cite{sier2023benchmark, qingqing2022multi}. However, for this study, we specifically utilize data from the Ouster lidar, the detailed specifications of which are provided in Table~\ref{tabs:ouster-table}.  Ouster lidar generates an incredibly dense point cloud along with various types of images. These images shown in Fig.~\ref{fig:ouster-lidar-img} include range images, signal images, and ambient images, each encoded with specific information: depth data, infrared intensity, and ambient light intensity, respectively. We particularly use range and signal images as they have been proven effective enough in the key point extraction~\cite{zhang2023lidar}.

The data sequences used for evaluation include indoor and outdoor environments. The outdoor environment is from the normal road and a forest, denoted as~\textit{Open road} and~\textit{Forest}, respectively. The indoor data includes a hall in a building and two rooms, denoted as~\textit{Hall (large)}, ~\textit{Lab space (hard)}, and~\textit{Lab space (easy)}, respectively. The~\textit{Forest} dataset we recorded was collected within a forested area, with a limited traversal distance of approximately 12 meters due to the constraints of our motion capture system. Thus, although the dataset was recorded outdoors, the environment is relatively confined.

\subsection{Proposed Point Cloud Sampling and its Evaluation}\label{subsec:evaluation_scheme}
\subsubsection{Overall Pipline}
The overall pipeline of the proposed algorithm is illustrated in Fig.~\ref{fig:overview}. Compared to our previous work~\cite{zhang2023lidar}, the proposed approach integrates DL-based super-resolution and colorization models, along with key point extractors utilizing the RGB three channels from colorized images within the processing pipeline. 

In the evaluation, we applied DL-based colorization approaches to the signal images, as our findings indicated that applying the models to the range images did not significantly improve keypoint extraction.

\subsubsection{Combination of Super Resolution and Colorization}
Given that super-resolution results in varying image sizes within the system, we conducted the evaluation using different combinations of images produced through super-resolution and colorization techniques. Table~\ref{tab:combinations} shows the exact combination we applied in the experiment. It is worth noting that our findings during the experiments indicate that the resolution size did not significantly affect the results of the effective key point extraction if it is above $2$.

\begin{table}[h]
    \centering
    \begin{tabular}{ll}
    \toprule
        \textbf{Name} & \textbf{Combinations} \\
    \midrule
        \textit{comb\_0}  &  [$rng$\; $sig$\; $sig_{2r}$]\, \\
        \textit{comb\_1}  &  [$rng$\; $sig$\; $sig_{c}$]\, \\
        \textit{comb\_2}  &  [$rng$\; $sig$\; $sig_{2r}^{c}$]\,\\
        \textit{comb\_3}  &  [$rng$\; $sig$\; $sig^{c}$\; $sig_{2r}$\; $sig_{2r}^c$]\,\\
        \textit{comb\_4}  &  [$rng$\; $rng_{2r}$\; $sig$\; $sig^{c}$\; $sig_{2r}$\; $sig_{2r}^{c}$]\,\\
        \textit{comb\_5}  &  [$rng$\; $rng_{2r}$\; $sig_{2r}$\; $sig_{2r}^{c}$]\,\\
        \textit{comb\_6}  &  [$rng$\; $rng_{2r}$\; $sig$\; $sig_{2r}^{c}$]\,\\
    \bottomrule
    \end{tabular}
    \caption{The different combinations of images from DL-based super-resolution and colorization are denoted as follows: $rng$ represents range images, $sig$ represents signal images, $_{2r}$ indicates a resolution size increased by a factor of two, and \(^{c}\) denotes the application of colorization.}
    \label{tab:combinations}
\end{table}


\subsubsection{Selection of DL-based Methods}
The super-resolution and colorization models utilized in this study are \textit{CARN} and \textit{DeOldify}, respectively. This selection is based on our analysis of both result quality and inference speed in the previous work~\cite{ha2024enhancing}. For keypoint extraction, we employed the \textit{Alike} model.

\subsubsection{Point Cloud Registration}
In this experiment, KISS-ICP~\footnote{https://github.com/PRBonn/kiss-icp.git} is employed as the method for calculating LO. Specifically, we disable the sampling functionality within the KISS-ICP code, as our approach focuses on sampling the point cloud using key point extractors.

\subsubsection{Evaluation Metrics}
We are particularly interested in understanding how the DL-based super-resolution and colorization techniques contribute to mitigating drift error and improving the accuracy of the LO system. To evaluate the accuracy of the LO, we calculated the translation and rotation errors using the tool evo~\footnote{https://github.com/MichaelGrupp/evo}. Additionally, we quantified the number of extracted point clouds to assess the effectiveness of these techniques.

\subsubsection{Evaluation Scheme}
The detailed evaluation scheme is outlined in Algorithm~\ref{alg:eva-test}. In this scheme, we systematically iterate through all image combinations with the corresponding DL-based approaches applied, listed in Table~\ref{tab:combinations}. We execute the full sequence of processes for each combination, including preprocessing, super-resolution, colorization, keypoint extraction, and LO calculation, followed by the computation of translation and rotation errors.

Given that lidar-generated images typically appear dark, a key objective during the preprocessing stage is to apply gamma compensation to enhance image brightness (Lines 6 - 13). 
Unlike range images, signal images exhibit highly uneven exposure across different regions. To address this issue, we first retain pixels with pixel values exceeding a predefined threshold, denoted as 
\(p_{thresh} = 240\) . For pixel values below this threshold, adaptive histogram equalization (CLAHE) is applied to enhance details in the darker regions of the signal image.
Following the preprocessing stage, super-resolution and colorization techniques were applied to the images as needed, depending on the specific combination of methods. Additionally, the key point extraction process was integrated into these enhanced images for further analysis.

To ensure greater robustness and consistency in key points across different image frames, we employed the Mutual Nearest Neighbor Matching(MNN) algorithm to match the key points extracted between the current and previous frames. We retained only the matched key points for subsequent processing.

After obtaining the robust key points, we apply the aforementioned LO approach and calculate the translation error and rotation error using the EVO tool.

\begin{algorithm}[h]
\DontPrintSemicolon
\SetNoFillComment
    \caption{The evaluation scheme}
    \label{alg:eva-test}
    \KwIn{\\
	\hspace{1em} Range image: $rng$ \\
        \hspace{1em} Signal image: $sig$ \\
        \hspace{1em} Point cloud: $pc$ \\
        \hspace{1em} Combinations in Table~\ref{tab:combinations}: $\textit{combs} = [\textit{comb\_i}]$, $i\in(0\sim6) $
    }
    \BlankLine
    \KwOut{\\
        \hspace{1em} Translation error: \(trans\_err\;(mean/rmse, unit:\,m)\) \\
        \hspace{1em} Rotation error: \(rot\_err\;(mean/rmse, unit:\,^\circ)\) \\
    }
    \BlankLine

    \tcc{Variable Declarations:}
$kps_{t}$ : Current Key points \;
$mkpts_{t}$ : matched Key points\;
$pc_{kp}$: Point cloud corresponded to key points \;
\textit{LOC} : LiDAR odometry calculation \;
\textit{GT} : Ground truth  \;
\textit{gamma} :  gamma transform adjusts the brightness  \;
\BlankLine
    \tcc{Image preprocess for brighter images}
    \SetKwFunction{gamma}{gamma}
    \SetKwFunction{histequal}{hist\_equalizer}
    \SetKwFunction{FPreproc}{img\_preprocess}
    \SetKwProg{Fn}{def}{:}{}
    \Fn{\FPreproc{$img$}}{
        \If{$img$ == $rng$}{
            $img_{prc}$ $\gets$ \gamma($img$)\;
        }
        \Else{
            \If{$pixel\_value < p\_thresh$}{     
                $img\_hist$ $\gets$ \histequal($img$)\;     
                $img\_prc$ $\gets$ \gamma($img\_hist$)\;
            }
        }
        \KwRet $img\_{prc}$; 
        }
    \BlankLine
    \tcc{Key point detect \& track}
    \SetKwFunction{FLOC}{LOC}
    \SetKwFunction{FEVO}{EVO}
    \SetKwFunction{FCombine}{Combine}
    \SetKwFunction{MnnMatcher}{MnnMatcher}
    \SetKwFunction{FTracker}{kp\_tracker}
    \SetKwFunction{FColorization}{colorization}
    \SetKwFunction{FSuperresolution}{super\_resolution}
    \SetKwFunction{FKeypointdet}{keypoint\_detect}
    \SetKwProg{Fn}{def}{:}{}
    \Fn{\FTracker{$kps_{t}$, $kps_{t-1}$}}{
            $matches$ $\gets$ \MnnMatcher{$kps_{t-1}$, $kps_{t}$}\;
            mkps $\gets$ $kps_{t}$[$matches$]\;
        \KwRet{mkps}\;
    }

    $rng$ = \FPreproc{$rng$}\; 
    \textit{$rng_{2r}$} $\gets$ \FSuperresolution{$rng$}  \\
    $sig$ = \FPreproc{$sig$}\;
    $sig^{c}$ $\gets$ \FColorization{$sig$}  \\
    $sig_{2r}$ $\gets$ \FSuperresolution{$sig$}  \\  
    $sig_{2r}^{c}$ $\gets$ \FColorization{$sig_{2r}$}  \\
    \tcc{Arrange and combine $sig$, $rng$, $rng_{2r}$, $sig_{2r}$,$sig_{2r}^{c}$,$sig^{c}$ to be \textit{comb} in table~\ref{tab:combinations}}
    \ForEach{$\textit{comb\_i}$ in $combs$}{
        \ForEach{$\textit{img}$ in $\textit{comb\_i} $}{

        $kps_{t}$ $\gets$ \FKeypointdet(\textit{img}) \;
        $mkpts_{t}$ $\gets$ \FTracker{$kps_{t}$, $kps_{t-1}$}\\
        $pc_{kp}^{i}$ $\gets$ pc[index[$mkpts_{t}$]] \\
}
        $pc_{kp}$ $\gets$ \FCombine{$pc_{kp}^{i}$} \\
        $Odom$ $\gets$ \FLOC{$pc_{kp}$} \\
        \textit{trans\_err}, \textit{rot\_err} $\gets$ \FEVO{$Odom$,GT} 
        \\
    }
    \vspace{-0.5em}
\end{algorithm}
\subsubsection{Hardware Information}
The evaluation was conducted using a Razer Blade 15 laptop by Ubuntu 22.04.4 LTS equipped with an Intel Core i7-12800H-20 processor, 16\,GB of RAM with a frequency of 4800\,MHz, and a GeForce RTX 3070 Ti GPU with 8\,GB of memory.

\section{Experimental Results}\label{sec:expert_resu}
To assess the efficacy of our proposed point cloud sampling approach, as outlined by various combinations in Table~\ref{tab:combinations}, we conducted an evaluation using ICP-based LO to calculate both its translation and rotation errors. The resulting errors, along with comparative results from prior studies~\cite{zhang2023lidar} across various scenarios, are summarized in Table~\ref{tab:error}. As the method proposed in the prior work outperforms the use of raw point clouds, a direct comparison with raw point cloud data is omitted.

The results indicate that the rotation errors associated with our method are generally lower than those reported in prior studies across all the data sequences from different environments.
Furthermore, our sampling method performs particularly well in more expansive environments, such as on ~\textit{Open road} and an~\textit{Hall(large)} datasets. However, in more confined spaces, such as~\textit{Forest}, ~\textit{Lab space(hard)}, and~\textit{Lab space (easy)}, our method exhibits slightly higher translation errors compared to existing approaches. 
Among the various combinations, ranging from comb\_0 to comb\_6, comb\_3 and comb\_4 exhibit the best performance in the majority of scenarios.

It is important to note that, unlike our previous work, the current methodology does not incorporate neighboring points surrounding the key points. This exclusion results in a significantly lower point density, as evidenced in Table~\ref{tab:points}, while still maintaining relatively competitive accuracy, as shown in Table~\ref{tab:error}.

\begin{table*}[t]
\resizebox{\textwidth}{!}{%
\begin{tabular}{@{}llcclll@{}}
\toprule
\multicolumn{2}{c}{\textbf{Combination}} & \textbf{Open road}                     & \textbf{Forest}                         & \textbf{Lab space (hard)}             & \textbf{Lab space (easy)}       & \textbf{Hall (large)}                  \\ 

\multicolumn{7}{c}{(Translation error (mean/rmse), rotation error (mean))} \\

\midrule

\multicolumn{2}{c}{comb\_0}  & (1.055/1.222, 1.782) & (0.086/0.102, 1.666) & (0.094/0.107, 1.185) & (0.083/0.095, 1.143) & (0.457/0.500, \textbf{0.903}) \\

\multicolumn{2}{c}{comb\_1}  & (0.605/0.731, 1.855) & (0.087/0.103, 1.663) & (0.120/0.136, 1.398) & (0.098/0.115, 1.216) & (0.485/0.532, 0.955) \\

\multicolumn{2}{c}{comb\_2} & (\textbf{0.357/0.409}, 1.808)  & (0.087/0.102, 1.668) & (0.126/0.149, 1.473) & (0.097/0.110, 1.186) & (0.456/0.500, 0.903) \\

\multicolumn{2}{c}{comb\_3} & (5.726/6.738, 1.957)  & (0.086/\textbf{0.102}, 1.666) & (0.045/0.050, \textbf{0.721}) & (0.032/0.036, \textbf{0.657}) & (\textbf{0.438/0.476}, 0.920) \\

\multicolumn{2}{c}{comb\_4} & (1.464/1.641, 1.806)  & (0.086/0.102, \textbf{1.662}) & (0.092/0.104, 1.235) & (0.080/0.095, 1.066) &    (1.175/1.326, 0.929) \\

\multicolumn{2}{c}{comb\_5} & (1.052/1.307, 1.807)  & (0.088/0.103, 1.672) & (0.101/0.114, 1.213) & (0.089/0.105, 1.118) &    (0.456/0.503, 0.922) \\ 

\multicolumn{2}{c}{comb\_6} & (0.492/0.595, 1.806)  & (0.086/0.103, 1.668) & (0.109/0.123, 1.319) & (0.093/0.107, 1.156) &    (0.452/0.498, 0.929) \\ 

\midrule

\multirow{3}{*}{\begin{tabular}[c]{@{}l@{}}Prior work~\cite{zhang2023lidar}\\ (\textit{rng\_sig)}\end{tabular}} &4\_7  & (0.817/0.952, 2.33) & (0.082/0.102, 7.88) & (0.039/0.046, 1.46) & (0.027/0.033, 0.98) & (0.583/0.660, 2.88) \\

&5\_5  & (2.176/2.410, \textbf{1.76}) & (0.108/0.203, 6.96) & (0.037/\textbf{0.043}, 1.35) & (0.027/0.032, 0.97) & (0.707/0.801, 2.66) \\

&7\_5  & (1.784/2.006, 2.30)  & \textbf{0.080}/0.102, 7.22) & (\textbf{0.033}/0.047, 1.59) & (\textbf{0.025/0.028}, 0.97) & (0.698/0.803, 3.11)  \\ \bottomrule
\end{tabular}
}
\caption{Comparison of translation and rotation errors across various combinations of DL-based super-resolution and colorization methods shown in Table~\ref{tab:combinations}, benchmarked against the results reported in prior work~\cite{zhang2023lidar}. In the table, 
\textit{sig} and \textit{rng} represent the size of neighboring point areas for the signal and range images, respectively, denoted as \textit{sig\_rng}.
}
\label{tab:error}
\end{table*}

\begin{table*}[]
\resizebox{\textwidth}{!}{%
\begin{tabular}{@{}llccccc@{}}
\toprule
\multicolumn{2}{c}{\textbf{Combination}} & \textbf{Open road} & \textbf{Forest} & \textbf{Lab space (hard)} & \textbf{Lab space (easy)} & \textbf{Hall (large)} \\ 

\multicolumn{7}{c}{(Number of Points (pts))} \\

\midrule

\multicolumn{2}{c}{comb\_0}     & 1149      & 1131   & 1613             & 1589             & 1492         \\
\multicolumn{2}{c}{comb\_1}     & 628       & 826    & 999              & 970              & 875          \\
\multicolumn{2}{c}{comb\_2}     & 787       & 820    & 1264             & 1229             & 1093         \\
\multicolumn{2}{c}{comb\_3}     & 1310      & 1167   & 1796             & 1270             & 1731         \\
\multicolumn{2}{c}{comb\_4}     & 1317      & 1170   & 2053             & 2031             & 1737         \\
\multicolumn{2}{c}{comb\_5}     & 1028      & 732    & 1587             & 1571             & 1322         \\ 
\multicolumn{2}{c}{comb\_6}     & 793       & 823    & 1396             & 1370             & 1099         \\ 

\midrule

\multirow{3}{*}{\begin{tabular}[c]{@{}l@{}}Prior work~\cite{zhang2023lidar}\\ (\textit{rng\_sig)}\end{tabular}} 

&4\_7          & 4784 & 11447  & 9518             & 9392            & 7094          \\

&5\_5          & 3183 & 7568   & 6446             & 6292            & 4783          \\

&7\_5          & 4756 & 11627  & 9469             & 9378            & 7078          \\ \bottomrule
\end{tabular}
}
\caption{Comparison of the number of points across various combinations of DL-based super-resolution and colorization methods shown in Table~\ref{tab:combinations}, benchmarked against the results reported in prior work~\cite{zhang2023lidar}. In the table, 
\textit{sig} and \textit{rng} represent the size of neighboring point areas for the signal and range images, respectively, denoted as \textit{sig\_rng}.}
\label{tab:points}
\end{table*}


\section{Conclusion and Future Work}\label{sec:conclusion}
This paper introduces a novel approach to point cloud sampling, specifically designed to mitigate drift error during the point cloud registration phase of LO. In contrast to our previous work, this study employs a DL-based super-resolution and colorization technique to enhance the key point extraction process for lidar-generated images. The proposed sampling method surpasses our previous work in terms of rotation error across most datasets and translation error in more open environments. However, it exhibits reduced accuracy in translation errors within more confined spaces.


Our findings suggest a promising strategy for reducing drift. In future work, this approach could be integrated into the entire LO process, such as by combining it with existing LO and SLAM methods, like Faster-LIO~\cite{bai2022faster}. Additionally, during our evaluation of the colorization and super-resolution models on lidar images, we observed that the existing models are primarily designed and trained on camera images rather than lidar images. While these models have shown significant improvements in the quality of lidar images, we believe their potential has yet to be fully realized. Therefore, future research could focus on developing and training colorization and super-resolution models specifically tailored for lidar images, which could further enhance the accuracy and performance of these techniques in the context of LO and SLAM systems.

\section*{Acknowledgment}
This research is supported by the Research Council of Finland's Digital Waters (DIWA) flagship (Grant No. 359247) and AeroPolis project (Grant No. 348480), as well as the DIWA Doctoral Training Pilot project funded by the Ministry of Education and Culture (Finland).

\vspace{0.6em}

\bibliographystyle{unsrt}
\bibliography{bibliography}

\end{document}